\documentclass{article}
\usepackage{spconf,amsmath,amssymb,graphicx,booktabs,multirow,xcolor,enumitem}
\usepackage[hidelinks]{hyperref}
\usepackage{pifont}
\newcommand{\cmark}{\ding{51}}
\newcommand{\xmark}{\ding{55}}

\title{TALK2AGENT: BENCHMARKING VOICE INTERFACES FOR TEXT AGENTS}

\name{
Terumi Chiba$^{\star}$
\qquad Guangzhi Sun$^{\dagger}$
\qquad Zheqi Yuan$^{\star}$
\qquad Chao Zhang$^{\star}$
}

\address{
$^{\star}$Department of Electronic Engineering, Tsinghua University \\
$^{\dagger}$University of Cambridge
}

\begin{document}
\ninept
\maketitle

\begin{abstract}
Large language model (LLM) computer-use agents are typically evaluated with clean written instructions, despite speech being an increasingly popular interface for interacting with such systems. Speech input introduces an additional failure point: transcription errors can alter task-critical entities, constraints, or targets before the agent begins reasoning, while conventional ASR metrics do not directly measure whether the information required for successful execution has been preserved. We introduce Talk2Agent, a benchmark for evaluating how effectively voice interfaces convey human-spoken instructions to LLM-based computer-use agents. Talk2Agent builds human-spoken versions of tasks from WildClawBench and OSWorld and evaluates a range of voice interfaces, including dedicated ASR models, audio-capable LLMs, contextual biasing, and LLM-based ontology repair. Because repeatedly executing long-horizon computer-use tasks is costly and stochastic, we further propose an execution-free, task-conditioned evaluation framework that projects the original task grader onto prompt-addressable intentions and measures how much task-relevant information is retained after the voice interface. On WildClawBench, Talk2Agent's execution-free native projection provides a practical, execution-grounded measure of voice-interface quality, correlating with downstream task completion and improving Pearson correlation by 0.246 over WER/CER on 32 hours of real human speech.

\end{abstract}


\section{Introduction}

Large language models (LLMs) are increasingly capable of performing agentic tasks, and benchmarks such as WebArena~\cite{zhou2024webarena} and OSWorld~\cite{osworld} assess them through functional outcomes in interactive environments. However, these benchmarks assume clean written user instructions. In practice, speech is becoming a natural interface for interacting with intelligent systems, introducing an additional and underexplored source of failure before an agent begins reasoning or acting. This is particularly important for computer-use tasks, where instructions often contain filenames, application names, paths, numbers, entities, and other task-critical details that are difficult to recognize reliably from speech. A transcription may appear largely correct while altering a single constraint or target that determines task success. Existing lexical or semantic ASR metrics therefore provide only an indirect measure of whether a spoken instruction remains useful to an agent, while evaluating every interface through full agent execution is costly and noisy, especially for long-horizon tasks.

We introduce Talk2Agent, a benchmark for evaluating how effectively voice interfaces convey human-spoken instructions to LLM-based computer-use agents. Talk2Agent builds spoken versions of tasks from WildClawBench (WCB)~\cite{wcb} and OSWorld~\cite{osworld}, covering both command-line and graphical computer interaction, and evaluates the resulting instructions with fixed downstream agents. This setup isolates the voice interface from the agent itself, and allows us to study how different interface designs preserve task-critical information. We evaluate both dedicated ASR systems and audio-capable LLMs as voice interfaces. Together with these interfaces, we investigate contextual biasing mechanisms that incorporate task-relevant environmental information, and LLM-based ontology repair that attempts to recover corrupted entities and terminology. In this way, Talk2Agent measures not only transcription quality, but the extent to which different voice interfaces preserve the information needed for successful agent execution.

Full execution, however, is expensive as a routine evaluation mechanism. Agent reliability can require repeated trials to characterize~\cite{yao2024taubench}; benchmarks such as WCB additionally involve long interaction trajectories and environment resets. We therefore additionally propose an execution-free evaluation framework that reuses the grading logic of the original text-agent benchmark. We decompose each task into prompt-addressable intentions and measure whether those intentions are retained after passing through the voice interface, weighting them according to their contribution to the original task grader. This produces task-conditioned retention measures that can diagnose interface errors without executing the downstream agent, while remaining grounded in what the benchmark ultimately rewards. We validate these measures against actual agent execution; on WCB, Native projection improves Pearson correlation coefficient (PCC) with task performance by an absolute 0.246 over WER/CER. Main contributions are summarized as follows:
\begin{itemize}[leftmargin=*]
  \item We propose Talk2Agent, an execution-grounded benchmark that isolates voice-interface quality for fixed LLM computer-use agents. Talk2Agent contains 31.82 hours of human-recorded instructions from WCB and OSWorld.
  \item A systematic study of voice interfaces for computer-use agents is performed, covering dedicated ASR models, audio-capable LLMs, contextual biasing, and LLM-based ontology repair.
  \item An execution-free, task-conditioned evaluation framework that projects an existing agent benchmark's grading logic onto instruction intentions, enabling efficient evaluation of voice interfaces without repeatedly executing expensive long-horizon tasks.
\end{itemize}

\section{Related Work}
Earlier spoken language understanding benchmarks evaluate intent and entity recognition~\cite{bastianelli2020slurp} and compositional semantic parsing~\cite{tomasello2023stop}. Talk2Agent instead targets the information needed for computer-use task execution.

Voice-agent benchmarks span spoken function calling and tool use~\cite{bfcl,voiceagentbench,pahwa2026audio2tool}, entity-sensitive task completion and conversation quality~\cite{eva}, and full-duplex voice agent tasks~\cite{tauvoice}. A closely related framework converts existing text tool-calling benchmarks into synthetic audio while retaining their tool schemas and labels~\cite{laskar2026textvoice}. These works primarily evaluate tool calling or end-to-end agents, where recognition, reasoning, and action jointly determine outcomes. In contrast, Talk2Agent uses real human speech for computer-use tasks and isolates the voice interface before a fixed downstream text agent while retaining task execution as the endpoint.

Beyond lexical WER/CER, ASR metrics assess semantic similarity~\cite{semdist,rugayan2023taskoriented,roy2021semantic,pipecat}, error severity~\cite{whetten2023severity}, downstream answer preservation~\cite{aer}, and atomic requirements~\cite{agenticasr}. SeMaScore~\cite{sasindran2024semascore} additionally weights aligned segments by their reference-context importance, while generative-LLM evaluation studies hypothesis preference and semantic error assessment~\cite{baneras2026asreval}. These approaches improve semantic or application-oriented assessment but do not directly encode the source computer-use benchmark's grading logic. Talk2Agent projects that grader onto prompt-addressable intentions, retaining its task-specific weights and non-additive interactions.

\section{Talk2Agent Benchmark}
\label{sec:benchmark}
\subsection{Benchmark construction}
Talk2Agent is constructed from two source benchmarks with complementary interaction modalities: WildClawBench (WCB) provides command-line tasks, and OSWorld provides graphical desktop tasks. We retain each source task and its original downstream evaluator so that an interface can be assessed by actual agent execution rather than transcript similarity alone. WCB contributes 60 formal tasks, excluding templates, and OSWorld contributes 351 English-language tasks.

Written agent prompts are not read verbatim. GPT-5.4 rewrites each prompt into a speakable script: prose becomes natural speech, lists and JSON are verbalized by sequence and fields rather than punctuation, and paths, filenames, and other exact tokens are spoken explicitly. The rewrite preserves execution-critical entities, constraints, literal symbols, and requested outputs; human review verifies that presentation changes do not alter task intent. Each WCB task is then recorded by ten Chinese--English proficient speakers (five women and five men). Each OSWorld task has five recordings; its participant pool contains 41 women and 57 men. Table~\ref{tab:data} summarizes the resulting 31.82 hours of human speech.

\begin{table}[t]
\centering\small
\caption{Talk2Agent speech collections. Coverage is tasks $\times$ recordings per task $=$ total recordings.}
\vspace{0.2cm}
\label{tab:data}
\begin{tabular}{@{}lcrl@{}}\toprule
Source & Coverage & Hours & Median [IQR] (s)\\\midrule
WCB & $60\times10=600$ & 24.84 & 113.66 [55.01--212.31]\\
OSWorld & $351\times5=1{,}755$ & 6.98 & 11.54 [7.00--18.28]\\\bottomrule
\end{tabular}
\end{table}

WCB recordings used smartphone or computer microphones in quiet indoor settings and were manually checked for completeness and intelligibility, with rerecording as needed. OSWorld recordings were collected through Prolific in 20-task batches under the same device and script-fidelity requirements.

\begin{figure*}[t]
  \centering
  \includegraphics[width=1\textwidth]{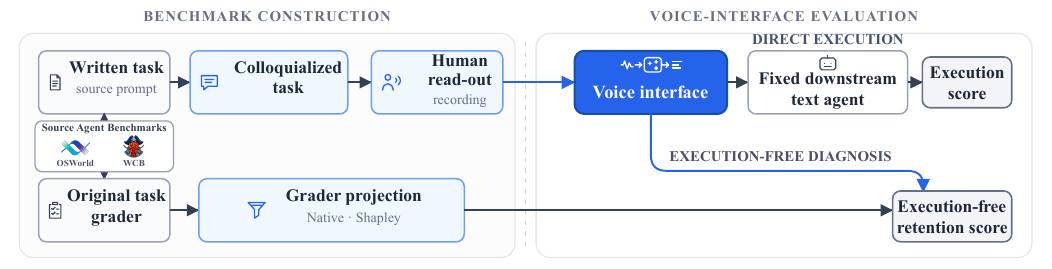}
  \vspace{-0.3cm}
  \caption{Talk2Agent dataset construction and evaluation. Written agent tasks are colloquialized to spoken forms and recorded by human speakers. Voice-interface outputs are evaluated either by a fixed downstream text agent and the source task grader or by execution-free projection of that grader onto retained task intentions.}
  \label{fig:framework}
\end{figure*}

Figure~\ref{fig:framework} shows the evaluation pipeline. Human speech is processed by a voice interface, whose text output is passed unchanged to a fixed downstream agent. The source benchmark evaluator then scores the resulting execution. This endpoint preserves the operational meaning of each task, but it also combines voice-interface information loss with agent capability, environment failures, and run stochasticity. Section~\ref{sec:execfree} therefore introduces a complementary execution-free diagnostic that evaluates the interface before downstream action.

\subsection{Voice-interface methods}
The benchmark compares vanilla ASR, audio-capable LLMs, and two agent-oriented interventions. \emph{Contextual biasing} supplies task-local terms from agent-visible workspace content, including GUI OCR, to improve recognition of rare and domain-specific entities~\cite{pundak2018deepcontext,sun2023contextual,huber2026newwords}. \emph{Ontology repair} applies Typeless-inspired LLM post-processing~\cite{typeless} to restore punctuation and technical terms only when supported by task-local context. We evaluate generative error correction~\cite{chen2023hyporadise} and contextual entity correction~\cite{song2023contextual,im2025deragec} as controlled benchmark interventions rather than propose a new correction algorithm. Original and colloquialized prompts serve as controls; model configurations are given in Sec.~\ref{sec:setup}.

\section{Execution-Free Evaluation}
\label{sec:execfree}
Direct execution is the benchmark's final endpoint, but it is expensive and noisy: observed score differences conflate voice-interface loss with downstream planning errors, limited agent capability, environment failures, and run-to-run variation. We therefore estimate the task-relevant information retained by an interface under an idealized downstream agent, while remaining grounded in the source benchmark's own grading logic.

For task $t$, we compile its evaluator into Boolean criteria $P_t=I_t\cup F_t$. $I_t$ contains the smallest prompt-addressable intentions, while $F_t$ contains fixed assessment conditions that affect grading but cannot be changed by the voice interface. Let $v_t(C)$ be the source-evaluator score when exactly criteria $C\subseteq P_t$ are satisfied. For a retained intention set $\mathcal R\subseteq I_t$, we condition on the fixed criteria and remove their baseline:
\begin{equation}
g_t(\mathcal R)=v_t(F_t\cup\mathcal R)-v_t(F_t).
\label{eq:conditional}
\end{equation}
This compilation preserves the evaluator's sums, products, gates, thresholds, caps, and floors; continuous criteria are represented by their unsatisfied and fully satisfied endpoints. Using the M\"obius representation of set functions~\cite{grabisch2000equivalent}, the conditioned game expands as $g_t(\mathcal R)=\sum_{T\subseteq\mathcal R}c_T$. Its exact Shapley values are $\phi_{ti}=\sum_{T\ni i}c_T/|T|$ and satisfy efficiency, $\sum_{i\in I_t}\phi_{ti}=g_t(I_t)-g_t(\emptyset)$~\cite{shapley}.

We report three complementary diagnostics. \emph{Lost intentions} counts the breadth of information loss. \emph{Native projection}
\begin{equation}
N_t(\mathcal R)=g_t(\mathcal R)/g_t(I_t)
\label{eq:native}
\end{equation}
substitutes retained intentions into the original grader and therefore preserves its joint logic. \emph{Shapley retention}
\begin{equation}
S_t(\mathcal R)=\sum_{i\in\mathcal R}\phi_{ti} \Big{/} \sum_{i\in I_t}\phi_{ti}
\label{eq:shapley}
\end{equation}
uses these values to additively allocate benchmark value to intentions. Native is an execution-oriented surrogate rather than a success probability; Shapley instead measures how much attributed task value survives, so neither score subsumes the other.

Each transcript intention is judged retained, changed, omitted, or unresolved; changed and omitted are treated as lost. For either $M_t\in\{N_t,S_t\}$, unresolved intentions $\mathcal U_t$ produce the exact interval
\begin{equation}
[M_{t,L},M_{t,U}]=\left[\min_{\mathcal Q\subseteq\mathcal U_t}M_t(\mathcal R_t\cup\mathcal Q),
\max_{\mathcal Q\subseteq\mathcal U_t}M_t(\mathcal R_t\cup\mathcal Q)\right].
\label{eq:interval}
\end{equation}
Midpoints are used only when a single value is required for aggregate analysis. Execution outcomes are never exposed during intention extraction, weighting, or retention judgment.

\section{Experimental Setup}
\label{sec:setup}
\subsection{Interface and agent configurations}
On WCB, we adopt ten interfaces cross Parakeet and Gemini recognition with contextual biasing and GPT-5.4 ontology repair, alongside original-text and colloquialized-text controls. Task-local bias terms are extracted by a simple prompt from agent-visible workspace crawls; scoring secrets are excluded. All WCB executions use GPT-5.6-Luna as the fixed downstream agent, with three runs per task--interface pair. To keep full execution tractable and control the acoustic input across interfaces, WCB execution uses TTS versions of the spoken prompts. Executing all ten human recordings per task would multiply the downstream-agent cost tenfold; these recordings are therefore more naturally analyzed using execution-free evaluation.
On OSWorld, the interface comparison includes GPT-4o, Whisper~\cite{radford2023whisper}, Parakeet, a Qwen recognizer, and Gemini end-to-end audio input. GPT-4o and Whisper ontology repair use GPT-5.4. Bias terms combine agent-visible workspace content with Apple Vision OCR in fast mode with language correction enabled. We use Gemini-3.5-flash and Qwen-3.5-9B as proprietary and open-source downstream agents, respectively, following the official OSWorld harness~\cite{osworld}.

\subsection{Execution-free validation protocol}
Before observing execution results, we audit each WCB prompt, programmatic grader, LLM-judge instruction, and aggregation rule. Requirements are split at the smallest independently scored criterion; reference-derived criteria remain linked to their parent requirement, and repetitions are not double counted. The 60 tasks yield 549 prompt-addressable intentions, with 2 to 46 per task (mean 9.15; median 6). GPT-5.6-Luna assigns retention labels using a fixed prompt without access to executions or scores. Unresolved labels account for $3.00\%$ of judgments; suspected errors are independently re-reviewed, and remaining uncertainty is propagated through Eq.~\ref{eq:interval}.

Correlation validation uses 50 non-Safety WCB tasks under ten matched conditions, for 500 task--interface observations. Safety tasks are excluded because missing an unsafe source intention can improve a safety-aligned execution score, reversing the ordinary retention-success relationship. We compare execution-free scores with direct execution and higher-is-better versions of WER/CER, SemDist, AER, Atomic Rubric, and Semantic WER. PCC is reported over 500 task--interface pairs, 50 task-type--interface pairs, and ten interface means. Task-grouped bootstrap intervals resample tasks while preserving all interfaces from each sampled task.

\section{Results}
\label{sec:results}
\subsection{Operational benchmark results}
Table~\ref{tab:wcbexec} reports the frozen 60-task WCB execution artifact, comprising ten matched interfaces and three runs per task--interface pair, with equal weighting across tasks and runs. All executions use GPT-5.6-Luna. These measurements are the operational endpoint, but they combine interface loss with downstream planning, environment failures, and run stochasticity. No intervention is uniformly beneficial in the full cohort. Parakeet changes little across settings. 

\begin{table}[h]
\centering\small
\caption{Results on Talk2Agent WCB split. Mean $\pm$ std. deviation reported across three round-level interface means; each round equally averages the 60 tasks. Voice rows use TTS audio.}
\vspace{0.2cm}
\label{tab:wcbexec}
\begin{tabular}{@{}lccl@{}}\toprule
Input / interface & Biasing & Ontology Repair & Execution (\%)\\\midrule
Original text &--&--&$40.75\pm1.67$\\
Spoken-form text &--&--&$36.39\pm3.56$\\
Parakeet &\xmark&\xmark&$24.20\pm2.76$\\
Parakeet &\xmark&\cmark&$25.32\pm1.13$\\
Parakeet &\cmark&\xmark&$25.28\pm3.39$\\
Parakeet &\cmark&\cmark&$25.03\pm2.95$\\
Gemini &\xmark&\xmark&$40.84\pm3.36$\\
Gemini &\xmark&\cmark&$37.32\pm1.33$\\
Gemini &\cmark&\xmark&$43.15\pm1.98$\\
Gemini &\cmark&\cmark&$38.68\pm2.22$\\\bottomrule
\end{tabular}
\end{table}

OSWorld supplies a distinct GUI operating point using human recordings (Table~\ref{tab:osworldexec}). Ontology repair raises the GPT-4o and Whisper execution means by $0.79$ and $0.70$ points with the Gemini agent, while Parakeet biasing changes them by only $+0.13$ with Gemini and $-0.11$ with Qwen. Thus execution effects depend on both the recognizer and downstream agent; OSWorld is not used in the WCB correlation analysis.

\begin{table}[h]
\centering\small
\caption{OSWorld execution from the frozen summary. Values are means and SDs over five speaker recordings per task}
\vspace{0.2cm}
\label{tab:osworldexec}
\begin{tabular}{@{}lccrr@{}}\toprule
Agent / input & Biasing & Ontology Repair & Mean & SD\\\midrule
\multicolumn{5}{@{}l}{\textit{Gemini agent}}\\
Text &--&--&61.76&0.37\\
GPT-4o &\xmark&\xmark&58.49&0.64\\
GPT-4o &\xmark&\cmark&59.28&1.09\\
Whisper &\xmark&\xmark&56.07&1.34\\
Whisper &\xmark&\cmark&56.77&1.50\\
Parakeet &\xmark&\xmark&56.26&1.59\\
Parakeet &\cmark&\xmark&56.39&1.22\\
Qwen recognizer &\xmark&\xmark&55.44&1.68\\
Gemini end-to-end &\xmark&\xmark&58.87&1.57\\\midrule
\multicolumn{5}{@{}l}{\textit{Qwen agent}}\\
Text &--&--&41.71&0.88\\
Parakeet &\xmark&\xmark&36.69&1.50\\
Parakeet &\cmark&\xmark&36.58&0.87\\\bottomrule
\end{tabular}
\end{table}

\subsection{Does execution-free scoring track execution?}
At the task--interface level, Native exceeds AER, the best generic metric, by $0.180$ (Table~\ref{tab:corr}). Shapley and Native exceed AER in 98.95\% and 99.89\% of task-grouped bootstrap replicates by PCC, and in 93.46\% and 99.13\% by Spearman correlation; they also rank first and second after task-type aggregation. At the interface-mean level, AER is numerically higher than Shapley ($0.916$ versus $0.902$), but not significantly so (95\% CI for $r_{\mathrm{AER}}-r_{\mathrm{Shapley}}$: $[-0.060,0.072]$), and is higher in only 53.35\% of replicates. Thus, task-conditioned metrics better track execution at the task--interface level, while remaining statistically comparable to the best generic metric at the interface-mean level.

\begin{table}[h]
\centering\small
\caption{PCC against WCB execution over 500 task--interface pairs, 50 task-type--interface aggregates, and ten interface means. Flat equally weights intentions; other metrics are in Sec.~\ref{sec:setup}.}
\vspace{0.2cm}
\label{tab:corr}
\resizebox{\columnwidth}{!}{%
\begin{tabular}{lrrr}\toprule
Metric & Task-interface & Type-interface & Interface\\\midrule
$1-$WER/CER & 0.220 & 0.176 & 0.514\\
$1-$SemDist & 0.218 & 0.178 & 0.606\\
$1-$AER & 0.286 & 0.297 & \textbf{0.916}\\
Atomic Rubric & 0.259 & 0.243 & 0.787\\
$1-$Semantic WER & 0.058 & 0.143 & 0.834\\
Flat retention & 0.349 & 0.441 & 0.895\\
Shapley retention & 0.399 & \textbf{0.476} & 0.902\\
Native projection & \textbf{0.466} & 0.475 & 0.870\\\bottomrule
\end{tabular}
}
\end{table}

\subsection{What does execution-free diagnosis reveal?}

\begin{table}[h]
\centering\small
\caption{Parakeet diagnosis on WCB. Targeted split comprises the 15 non-Safety tasks in the preselected 16-task workspace-intensive probe. Full comprises all 50 non-Safety WCB tasks. Bold indicates improvements from ontology repair under matched biasing. All values are percentages.}
\vspace{0.2cm}
\label{tab:diagnosis}
\begin{tabular}{@{}lllrrr@{}}\toprule
Cohort & Biasing & Ont. fix & Shapley & Native & Execution\\\midrule
Targeted & \xmark & \xmark &39.93&14.00&40.60\\
($n=15$) & \xmark & \cmark &\textbf{69.33}&\textbf{34.00}&\textbf{52.45}\\
& \cmark & \xmark &41.30&14.67&33.97\\
& \cmark & \cmark &\textbf{59.25}&\textbf{33.67}&\textbf{50.06}\\\midrule
Full & \xmark & \xmark &33.39&7.73&22.64\\
($n=50$) & \xmark & \cmark &\textbf{54.84}&\textbf{23.24}&22.68\\
& \cmark & \xmark &33.44&7.33&22.38\\
& \cmark & \cmark &\textbf{46.30}&\textbf{16.33}&22.64\\\bottomrule
\end{tabular}
\end{table}

\begin{table}[t]
\centering\small
\caption{OSWorld execution-free transfer, mean $\pm$ SD across five paired Whisper takes. Improvements are percentage points.}
\vspace{0.2cm}
\label{tab:osworldhuman}
\setlength{\tabcolsep}{3pt}
\begin{tabular}{@{}lrrr@{}}\toprule
Measure & Raw & +Ont. & $\Delta$\\\midrule
Whisper & $54.93\pm1.18$ & $62.89\pm2.05$ & $+7.96\pm1.52$\\\bottomrule
\end{tabular}
\end{table}

We next test a mechanism-specific operating point. Before observing intervention outcomes, we selected the 15 WCB tasks with the most task-relevant workspace content because bias lists are extracted from that workspace. As shown in Table\ref{tab:diagnosis}, Ontology repair improves Parakeet execution by $11.85\%$ without biasing and $16.09\%$ with biasing; biasing itself has no consistent benefit. The same direction appears in Shapley and Native projection. Across all 50 tasks, ontology repair still recovers substantial interface-level value and joint grader logic, while end-to-end execution changes by less than $0.3$\%. Thus the targeted probe exposes the operational effect where the intervention is relevant, whereas full-cohort retention isolates an interface improvement obscured by downstream execution variability.

Moreover, Shapley retention and Native projection provide complementary views: Shapley attributes grader value to individual intentions, whereas Native evaluates them under the grader's original composition. On the full cohort without biasing, ontology repair yields Shapley retention of $54.84\%$ but Native projection of $23.24\%$, suggesting that many individually valuable requirements are restored but some joint requirements are not satisfied.

Finally, the execution-free evaluation transfers from WCB's command-line tasks to OSWorld's graphical tasks. Across five paired Whisper takes, ontology repair improves the intent accuracy with $7.96$-point gain (Table~\ref{tab:osworldhuman}). The smaller directionally consistent execution gains in Table~\ref{tab:osworldexec} reflect the additional downstream-agent factors absent from interface-level retention.

\section{Conclusion}
We presented Talk2Agent, a human-spoken benchmark and an execution-free evaluation framework for voice interfaces to LLM agents. By projecting source-benchmark graders onto prompt-addressable intentions, its Native and Shapley measures complementary diagnosis of task-relevant information loss without repeatedly executing the downstream agent, while remaining operationally grounded through execution-based validation. The study also exposes several open challenges. Long, written prompts do not necessarily reflect natural voice use. Speakers may omit, paraphrase, or interactively resolve URLs, file paths, code, and other structured content rather than verbalize them in full. Moreover, bias lists constructed from workspace did not consistently improve agentic task execution, leaving open how relevant context should be selected and divided between interface-level recognition and agent-level error recovery.

\clearpage

\bibliographystyle{IEEEbib}
\bibliography{talk2agent}
\end{document}